\pdfoutput=1

\documentclass[11pt]{article}
\usepackage[dvipsnames,usenames]{xcolor}

\usepackage{acl}

\usepackage{times}
\usepackage{latexsym}

\usepackage[T1]{fontenc}

\usepackage[utf8]{inputenc}

\usepackage{microtype}

\usepackage{inconsolata}

\usepackage{booktabs}
\usepackage{algorithm} 
\usepackage{algpseudocode}
\usepackage{graphicx}
\usepackage{amssymb}
\usepackage{soul}
\usepackage{multirow}

\newcommand{\kmy}[1]{\textcolor{magenta}{$_{min}$[#1]}}

\newcommand{\feedback}[0]{\textit{ISQA}}

\newcommand{\fbmodel}{{\mathcal{F}}}
\newcommand{\summodel}{{\mathcal{G}}}

\title{\feedback{}: Informative Factuality Feedback for Scientific Summarization}

\author{Zekai Li\\
  National University of Singapore \\
  \texttt{lizekai@u.nus.edu} \\\And
  Yanxia Qin \\
  National University of Singapore \\
  \texttt{qinyx@comp.nus.edu.sg} \AND 
  Qian Liu \\
  Sea AI Lab \\
  \texttt{liuqian@sea.com} \\\And
  Min-Yen Kan \\
  National University of Singapore\\
  \texttt{kanmy@comp.nus.edu.sg}}
  
\begin{document}
\maketitle
\begin{abstract}


We propose \textit{Iterative Facuality Refining on \textbf{I}nformative \textbf{S}cientific \textbf{Q}uestion--\textbf{A}nswering (\textbf{ISQA}) feedback}\footnote{Code is available at \url{https://github.com/lizekai-richard/isqa}}, a method following human learning theories that employs model-generated feedback consisting of both positive and negative information. 
Through iterative refining of summaries, it probes for the underlying rationale of statements to enhance the factuality of scientific summarization. \feedback{} does this in a fine-grained manner by asking a summarization agent to reinforce validated statements in positive feedback and fix incorrect ones in negative feedback. 
Our findings demonstrate that the \feedback{} feedback mechanism significantly improves the factuality of various open-source LLMs on the summarization task, as evaluated across multiple scientific datasets. 
\end{abstract}

\section{Introduction}

Large language model (LLM) based scientific summarization toolkits facilitate quicker screening of a large number of daily emerged scientific documents by providing concise and fluent summaries. However, the hallucination nature of generative LLMs introduces a fundamental flaw in the system-generated summaries: inconsistency with the original scientific documents~\cite{lin-etal-2022-truthfulqa, zhang2023language, min-etal-2023-factscore, chen-etal-2021-factuality-checkers, krishna-etal-2023-longeval}. Despite their fluency, these nonfactual summaries generated by LLMs can mislead scientists, due to the unreliable nature of the information provided. 

Recent studies on refining LLM-generated content through feedback mechanisms have gained significant attention \cite{liu-etal-2023-improving, roit-etal-2023-factually, chen2023iterative,pan2023automatically}. 
These studies enhance the output of an LLM by providing feedback on its previous outputs, thereby prompting the generation of improved content. Feedback mechanisms offer an effective and efficient post-hoc method to enhance LLM-generated content without the need for fine-tuning the LLM itself. In this work, we continue this line of work by exploring how feedback can be utilized to improve the factual consistency of LLM-generated scientific summaries.

\begin{figure}[t]
    \centering
    \includegraphics[width=0.5\textwidth]{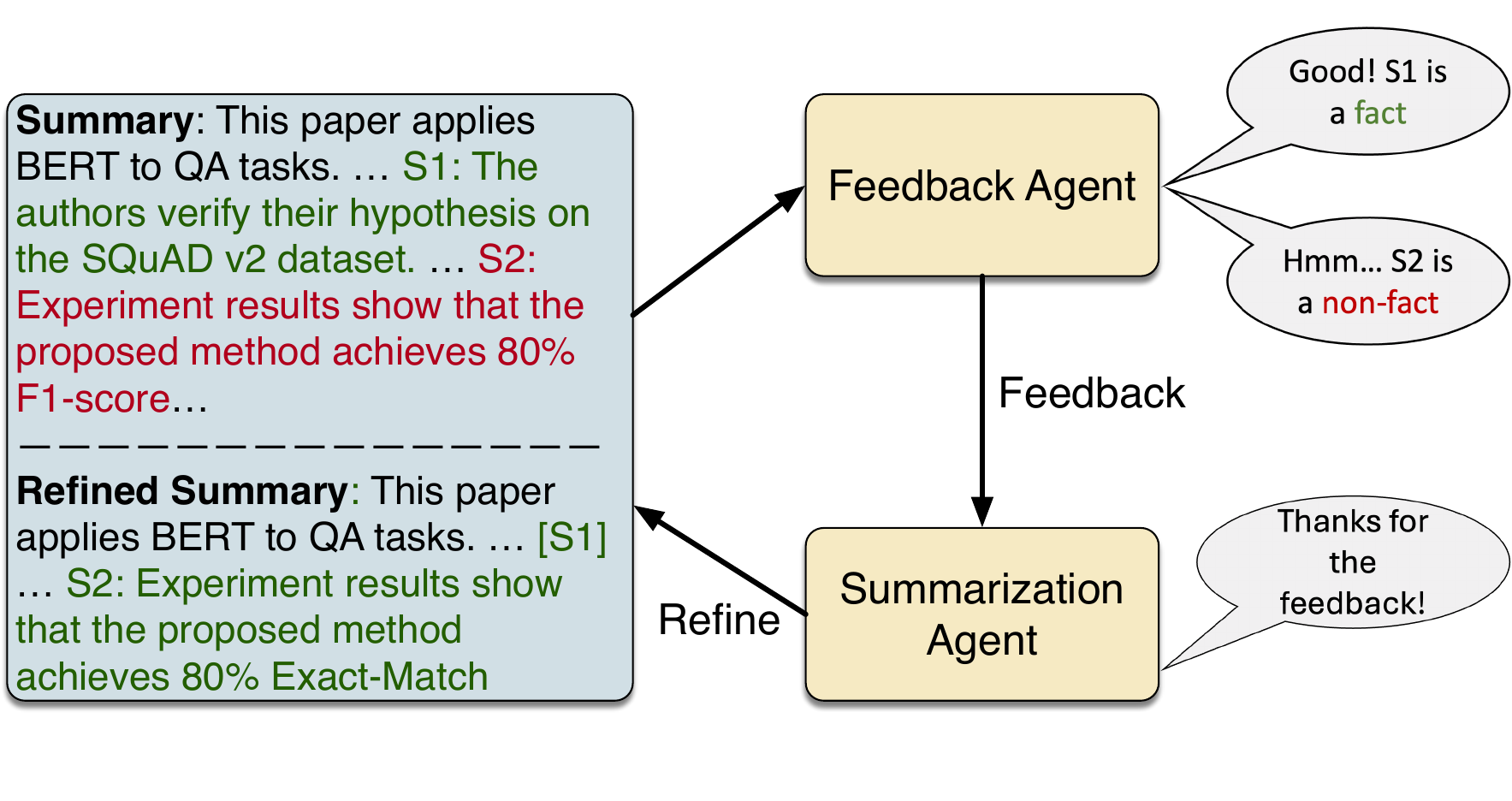}
    \caption{Illustration of improving factuality of summarization utilizing both positive (fact) and negative (non-fact) feedback. The facts and non-facts are highlighted in \textcolor{ForestGreen}{green} and \textcolor{Maroon}{red}, respectively. The summarization agent refines its output by remembering the facts and correcting non-facts.}
    \label{fig:intro}
\end{figure}



Previous research has explored various sources of feedback, including human annotations~\cite{liu-etal-2023-improving}, model-generated feedback~\cite{roit-etal-2023-factually, chen2023iterative}, external tools~\cite{nathani-etal-2023-maf}, and self-correction mechanisms \cite{madaan2023selfrefine}. They largely emphasize the use of positive feedback to encourage desired outcomes, such as accurate reasoning paths. Yet, an essential aspect of the learning process, as highlighted by human learning theories like operant conditioning\footnote{\url{https://en.wikipedia.org/wiki/Operant\_conditioning}}, is the strategic use of negative feedback. This type of feedback is instrumental in discouraging undesired behaviors, serving as a complementary role to positive feedback, yet has been overlooked in existing literature. In this work, we recognize the value of both positive and negative feedback and aim to combine them together for factuality enhancement. As depicted in Figure \ref{fig:intro}, we expect a feedback agent to inspect a summary output and produce both positive feedback and negative feedback accordingly, which will be provided to the summarization agent to refine the summary.



We propose to use model-generated positive feedback---specifically, facts\footnote{In this work, facts are defined to be factually consistent content in the summary. Similar definition to non-facts.} within summaries---to reinforce the generation of factual content by a summarization LLM, and use negative feedback---non-facts---to punish the generation of non-factual content. 
To distinguish between factual and non-factual information in the generated summaries for the purpose of providing accurate feedback, we employ a Question-Answering (QA) based approach.
The intuition is that by asking questions about the summary and comparing the predicted answers with the ground-truth answer, we can infer the factuality of the evidence sentence in the summary that supports the answer.
It is classified as a fact if the answer matches the ground truth, and otherwise non-fact.
We refer to the proposed feedback mechanism as \textbf{I}nformative \textbf{S}cientific \textbf{Q}uestion-\textbf{A}nswering (\textbf{ISQA}) feedback. 

In response to two distinct types of informative feedback, we use different prompts for the summarization agent to process each feedback type appropriately. Echoing previous studies \cite{madaan2023selfrefine, nathani-etal-2023-maf, chen2023iterative}, we leverage an iterative refinement approach to gradually improve the summary. 
Through extensive experimentation with three open-source LLMs as summarization agents, our results affirm that the \feedback{} enables LLMs to significantly improve the factuality of summaries over two prominent scientific document datasets. Notably, the performance improvement of \feedback{} surpasses that achieved through the application of both generic feedback and prompt engineering. Moreover, our findings reveal that relatively small language models, with the \feedback{} mechanism, can generate high-quality feedback comparable to those by large language models (e.g., GPT-3.5), which results in high-factuality summary output.

\section{Related Work}
\paragraph{Factuality of Abstractive Summarization} 
Recent advanced automatic text summarization systems, especially LLM-based, have demonstrated a strong capability of generating fluent, controllable, and informative summaries\cite{aharoni-etal-2023-multilingual, bhattacharjee-etal-2023-crosssum, chen-etal-2023-unisumm, he-etal-2023-z}. Despite performing well, these model-based summarization systems still confront certain issues that remain unsolved, one of which is generating factually inconsistent summaries \cite{balachandran-etal-2022-correcting, tang-etal-2023-understanding, liu-etal-2023-improving}. According to~\citet{chen-etal-2021-factuality-checkers}, a summary of high factuality is supposed to be supported well by the source document without using any external knowledge. By contrast, poor factuality indicates the existence of factual errors, which are statements that contradict the source document or are not directly stated, heavily implied, or logically entailed by the source document~\cite{krishna-etal-2023-longeval}. In the context of scientific summarization, it is crucial to ensure the reliability of system-generated summaries for the research society to access and review. Therefore, in addressing this pressing issue, we introduce informative factuality feedback as a means to effectively refine the summary generation, thereby achieving an overall enhancement in factuality.


\begin{figure*}[t]
\centering
\includegraphics[width=0.9\textwidth,height=0.45\textwidth]{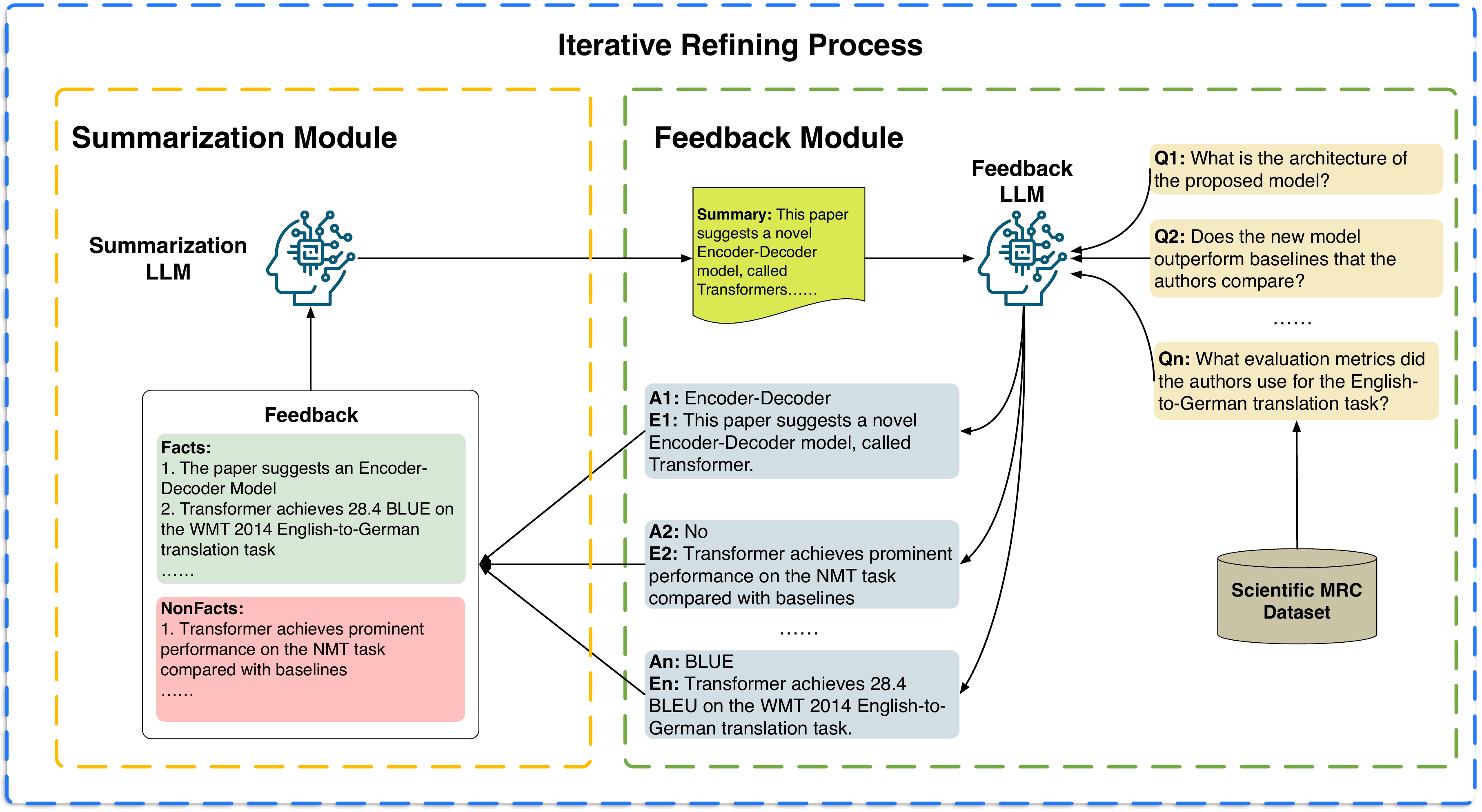}
\caption{The framework of the proposed method. The feedback LLM is trained to examine the factuality of the summary and then provide informative feedback by asking insightful questions. The summarization LLM then incorporates the feedback to produce a summary with improved factuality. The interaction between two modules is an iterative refining process.}
\label{fig:refine}
\end{figure*}


\paragraph{Automated Feedback for LLMs} Recent progress in rectifying LLMs' unsatisfactory behaviors mainly focuses on asking LLMs to learn from automated feedback~\cite{pan2023automatically,learningfrommistake2023}. 
There are also two sources of automated feedback: external tools and LLM. External feedback is usually obtained by employing external tools or knowledge sources such as search engines~\cite{gou2024critic}. Different external tools can also be utilized together.\citet{nathani-etal-2023-maf} argues the lack of error-specific feedback and proposes multi-aspect feedback, which seeks help from external tools based on different types of errors. In contrast, LLM-generated feedback comes from LLM critiquing its own output and corrects it accordingly~\cite{du2023improving,welleck2022generating,huang2022large}. Self-Refine~\cite{madaan2023selfrefine} proposes an iterative prompt refinement approach and applies it to various reasoning tasks where remarkable improvements have been witnessed. \citet{liu-etal-2023-improving} presents a method leveraging natural language feedback to enhance the factual consistency of summaries through a self-constructed dataset with human demonstrations. In this work, we adopt LLM-generated feedback in the natural language format. However, we don't rely on LLMs' self-correction ability which has only been witnessed in the large commercial models, e.g. GPT4~\cite{madaan2023selfrefine}. Instead, our \feedback{} feedback serves as guidance on factuality enhancement for LLMs to gradually improve.


\section{Method}
As illustrated in Figure \ref{fig:refine}, the proposed method contains a summarization module and a feedback module, which interact iteratively. The feedback module is designed to provide informative feedback on the system-generated summary to enhance its factuality. Subsequently, the summarization module utilizes this feedback, along with the source document, to produce an improved summary.

\subsection{Feedback Module}
\label{sec:fm}
The feedback module plays a critical role akin to that of a doctor, tasked with examining the patient and offering diagnoses, which can be done by asking the patient a series of symptom-related questions. In our scenario, the summarization module is the patient who contracts a low factuality "disease". Through \feedback{}, we enable the feedback module, as the doctor, to provide feedback on factuality directly.



\paragraph{Informative Scientific Question--Answering (\feedback{}) Feedback}
A diagnosis made by a doctor should align with the patient's condition, and similarly, feedback provided by the feedback module should correspond appropriately. In the context of enhancing the factuality of summarization, we posit that factual and non-factual elements in the generated summaries could serve as the most direct and effective content for feedback. Thus, we introduce \feedback{} which realizes the above functionality via the following two design features:


1) \textit{Evidence-seeking QA}: As mentioned earlier, the supporting evidence that underpins the answer is the foundational source of the \feedback{} feedback. To acquire the evidence, we introduce an \textit{evidence-seeking QA} task where the model is designed to elucidate its reasoning process through the provision of the evidence sentence, in addition to the answer prediction in the conventional QA tasks. The evidence sentences extracted are then categorically processed into facts and non-facts, which form the basis of the positive and negative aspects of the \feedback{} feedback, respectively. By adopting this innovative approach, we empower the feedback module to effectively analyze the factuality of summary content, thereby facilitating the generation of informative feedback for factuality refinement. 

2) \textit{Employ Scientific Questions}: The \feedback{} feedback is constructed to offer comprehensive and domain-specific insights, meaning the feedback content should encompass information of varying depths and possess substantial scientific value. The goal is accomplished through the employment of high-quality scientific questions, especially human-annotated datasets. Unlike questions from other sources, such as those generated by language models, human-labeled data guarantees that the questions meet the expectations of being domain-specific and encompass a range of difficulty levels.

Formally, the \feedback{} mechanism is defined as follows: given a set of scientific questions paired with its context $\{(q_1, c_1), (q_2, c_2), \cdots, (q_m, c_m)\}$ for a scientific paper $x$, the feedback model $\fbmodel$ is expected to output both the answer $a$ and its supporting evidence $e$,
\begin{equation}
\label{eq:fb}
    a, e = \fbmodel(c, q).
\end{equation}
Based on the answer correctness, evidence $e$ is categorized into either positive or negative feedback. Refer to Algorithm\ref{algo:refine} line 7-13 for details.

\paragraph{Feedback LLM}
Recent studies~\cite{madaan2023selfrefine,nathani-etal-2023-maf} show that vanilla open-source LLMs, such as Vicuna~\cite{vicuna2023}, lack the capability to critique or correct their outputs\footnote{This limitation also prevents vanilla LLMs from simultaneously acting as both the summarization LLM and the feedback LLM. Consequently, we employ distinct models for these tasks.}. 
Thus, we propose to fine-tune the vanilla LLMs with \feedback{} as the specific task, to specialize in summary inspection and the generation of informative factuality feedback. Please refer to Appendix~\ref{sec:fbft} for details of feedback LLM fine-tuning. It's worth noting that we still rely on the vanilla feedback LLM for generating fluent and coherent texts and following instructions\footnote{This requires the feedback LLM to be instruction-tuned.}. We argue that the feedback model is capable of offering both insightful positive and negative feedback, provided it can accurately match the supporting evidence with the answers generated after being fine-tuned.

\subsection{Summarization Module}
The main component of the summarization module is the summarization LLM, denoted as $\summodel$, which is designed to process source documents alongside feedback and output a summary. To accommodate these inputs, we craft both summary-generation and act-on-feedback prompts\footnote{Natural language prompts are adopted in this work for better interpretability.}. In designing the summarization prompt, we adhere to common practices. The act-on-feedback prompt is crafted according to human learning theories: facts are expected to be memorized, while non-facts are expected to be rectified or forgotten. Given the evolving nature of feedback prompts, we employ instruction-tuned LLMs to bolster their comprehension of instructions. We argue that the LLM can enhance its output by encouraging fact generation and discouraging non-facts under the guidance of feedback prompts. We present the following template for summarization prompts\footnote{Efforts are made to keep prompts succinct, with the maximum number of facts and non-facts capped at eight to adhere to the input context length limitations of open-source LLMs}:

\noindent \texttt{Below is a scientific paper and a set of facts and non-facts in the paper. Please write a summary by memorizing facts and rectifying non-facts.\\
Facts: \{$sent_1, sent_2, \cdots, sent_n$\}\\
Non-Facts: \{$sent_1, sent_2, \cdots, sent_n$\}\\
Paper: ``This is the body text of a scientific paper''.}
\subsection{Iterative Refining Process}
In alignment with theories of human learning through feedback and insights from prior research, we adopt an iterative refinement strategy. Algorithm \ref{algo:refine} shows the refinement process for a single scientific paper.

\paragraph{Initialization} The inputs include the summarization LLM $\summodel$, feedback LLM $\fbmodel$, initial prompt for summarization $p_0$, scientific paper $x$ and corresponding QA pairs $D$ from the dataset, the maximum number of refining iterations $N$, and initially empty sets for facts $F$ and non-facts $\overline{F}$.

\begin{algorithm}[t]
\small
	\caption{Iterative Refinement Process} 
        \renewcommand{\algorithmicrequire}{\textbf{Input:}}
	\renewcommand{\algorithmicensure}{\textbf{Output:}}
	\begin{algorithmic}[1]
            \State {$\textbf{Input:}\ \summodel, \fbmodel, p_0, D, N$}
            \State {Initialize $F=\varnothing, \overline{F}=\varnothing$} 
		\For {$t = 1, 2, \ldots N$}
                \State {$B \sim D ~~~~~~~~~~~~~~~~~~\leftarrow$ \textit{select a random QA batch}}
			\State {$s_t \leftarrow \summodel(p_{t-1})$}
                \For {$q_i, a^*_i$ \textbf{in} $B$} ~~~~~~~~~~~~$\leftarrow$ \textit{batched refinement}
                    \State {$a_i, e_i \leftarrow \fbmodel(s_t, q_i)$}
                    \State {$simScore \leftarrow F1(a_i, a^*_i)$}
                    \If {$simScore \geq \epsilon$}
                    \State {add $e_i$ to $F$}
                    \Else 
                    \State {add $e_i$ to $\overline{F}$}
                    \EndIf
                \EndFor
            \State{$p_t$ $\leftarrow$ update $p_{t-1}$ with $F$ and $\overline{F}$}
		\EndFor
	\end{algorithmic} 
 \label{algo:refine}
\end{algorithm}


\paragraph{Batched Feedback Retrieval} To mitigate the potentially destabilizing effects of noisy questions---such as those outside the scope or of poor quality---we adopt a batched strategy, as opposed to generating feedback after a single QA interaction. 
Specifically, we update the summary after processing a mini-batch of QA pairs, which are randomly selected from the complete set of QA pairs $D$. 
For each QA pair ($q_i, a^*_i$) in the batch, the feedback model addresses the question $q_i$ using the current summary $s_t$ and produces a predicted answer $a_i$ along with supporting evidence $e_i$. We then calculate the text similarity (token-level F1) score between the predicted answer $a_i$ and the ground truth answer $a^*_i$. Based on this score, $e_i$ is classified as a fact in $F$ if the score exceeds a predefined threshold $\epsilon$; otherwise, as a non-fact in $\overline{F}$.

\section{Experiment}
\subsection{Datasets}
\paragraph{SciMRC~\cite{zhang2023scimrc}}SciMRC is a scientific machine reading comprehension dataset with questions across three difficulty levels: beginner, student, and expert. It includes 741 papers with a total of 6,057 QA pairs, distributed as 3,306 for beginners, 1,800 for students, and 951 for expert levels. We only use the training and validation sets for ground truth answers, consisting of 592 papers and 4873 QA pairs. Note that although factuality evaluation doesn't require summary labels, we take the abstract as the reference summary for each paper to evaluate the coherence of generated summaries(See Section~\ref{sec:main_results}).

\paragraph{QASPER~\cite{dasigi-etal-2021-dataset}} QASPER is a dataset comprised of 5,049 questions over 1,585 Natural Language Processing (NLP) papers. Questions are provided by NLP practitioners while answers are extracted by another set of practitioners who also provide corresponding supporting facts. For the same reason, we only use the training and validation sets, with 1169 papers in total. Reference summaries are extracted in the same way.

\subsection{Baseline Models}
\paragraph{Summarization Model} To demonstrate the efficacy of our proposed \feedback{} feedback mechanism, we utilize three open-source, instruction-tuned mainstream LLMs as our summarization baseline models: \textit{Llama2-7b-chat}~\cite{touvron2023llama}, \textit{Vicuna-7b}~\cite{vicuna2023}, and \textit{Flan-T5-11b}~\cite{chung2022scaling}. It's important to note that these LLMs operate in a zero-shot manner for summarization tasks, meaning they will not undergo fine-tuning, even when feedback is provided. Please refer to Appendix~\ref{sec:bpp} for details of baseline summary generation prompts.

\paragraph{Feedback Model} For the feedback model, we have empirically selected \textit{Vicuna-7b} as the base model, which is further fine-tuned to excel in summary examination and providing constructive feedback. To benchmark against high-performing commercial solutions, we include GPT-3.5 as a baseline model\footnote{GPT-3.5 is chosen over GPT-4 due to cost considerations.}. The comparative analysis with GPT-3.5 is detailed in Section \ref{sec:gpt}.

\begin{table*}[t]
\centering
\small
\begin{tabular}{clcccccc}
\toprule
\multicolumn{2}{c}{\multirow{3}{*}{\begin{tabular}[c]{@{}c@{}} Models\end{tabular}}}                 & \multicolumn{3}{c}{SciMRC}                                                                                          & \multicolumn{3}{c}{QASPER}                                              \\ \cline{3-8} 
\multicolumn{2}{c}{}                                        & \multicolumn{2}{c}{\textbf{Factuality}}                                       & \textbf{Quality}                           & \multicolumn{2}{c}{\textbf{Factuality}} & \multicolumn{1}{c}{\textbf{Quality}} \\ \cline{3-8} 
\multicolumn{2}{c}{}                                        & \multicolumn{1}{l}{QAGS $\uparrow$} & \multicolumn{1}{l}{QuestEval$\uparrow$} & \multicolumn{1}{l}{ROUGE$\uparrow$} & QAGS $\uparrow$   & QuestEval$\uparrow$ & ROUGE$\uparrow$                         \\ 
\midrule
\multirow{2}{*}{\textit{Llama2-7b-chat}}  & \multicolumn{1}{c}{\textit{w/o \feedback{}}} & 54.53\%                             & 28.17\%                                 & 28.04\%                             & 43.48\%           & 27.83\%             & 25.54\%                       \\
                         & \textit{w/ \feedback{}}                      & \textbf{61.41\%}                    & \textbf{33.62\%}                        & \textbf{30.98\%}                             & \textbf{48.82\%}  & \textbf{28.67\%}    & \textbf{26.78\%}                       \\ \hline
\multirow{2}{*}{\textit{Vicuna-7b}}  & \textit{w/o \feedback{}}                     & 50.38\%                             & 29.07\%                                 & 30.21\%                             & 49.60\%           & 27.32\%             & 25.81\%                       \\
                         & \textit{w/ \feedback{}}                      & \textbf{58.52\%}                    & \textbf{32.96\%}                        & \textbf{31.77\%}                             & \textbf{56.19\%}  & \textbf{32.82\%}    & \textbf{27.19\%}                       \\ \hline
\multirow{2}{*}{\textit{FLAN-T5-11b}} & \textit{w/o \feedback{}}                     & 57.14\%                             & 28.90\%                                 & 35.94\%                             & 54.47\%           & 26.46\%             & 26.04\%                       \\
                         & \textit{w/ \feedback{}}                      & \textbf{61.91\%}                    & \textbf{32.42\%}                        & 35.19\%                             & \textbf{58.33\%}  & \textbf{29.63\%}    & \textbf{26.77\%}                       \\ 
                         \bottomrule
\end{tabular}
\caption{Results of three summarization LLMs with/without our proposed \feedback{} feedback on two datasets, SciMRC and QASPER. The baseline (w/o \feedback{}) results refer to the best among different baseline prompts. We evaluate the summarization in two aspects: factuality and text quality, with the main focus on the former. Bolded figures indicate improved results when compared to baseline (w/o \feedback{}). }
\label{tab:main}
\end{table*}

\subsection{Evaluation Metrics}
\paragraph{QAGS~\cite{wang2020asking}}
QAGS is a QA-based metric to evaluate the factuality of model-generated content. It operates on the principle that asking questions about a summary and its source and then comparing the answers can reveal factual inconsistencies. If the summary is factually consistent with the source, similar answers are expected for the same questions asked of both. This method has shown substantially higher correlations with human judgments of factual consistency compared to other automatic evaluation metrics.

\paragraph{QuestEval~\cite{scialom-etal-2021-questeval}}
QuestEval introduces another QA-based summarization factuality evaluation metric. Building upon previous precision-oriented QA-based metrics, including QAGS~\cite{wang2020asking}, QuestEval integrates precision and recall~\cite{scialom-etal-2019-answers} where the precision measures the consistency of generated summaries by token-level F1-scores of predicted answers, and the recall measures how much important information that generated summaries contain by query weighting\footnote{QuestEval also demonstrated its capability of better aligning human judgments in determining the factual consistency of summarization.}.

\paragraph{ROUGE} 
The ROUGE~\cite{lin-2004-ROUGE} scores measure the overlap of n-grams between machine-generated texts and reference texts. Higher ROUGE scores suggest closer similarity to human-written summaries, making it a standard text quality evaluation metric. We use the average of ROUGE-1, ROUGE-2, and ROUGE-L, denoted as ROUGE.

\subsection{Main Results}
\label{sec:main_results}
As illustrated in Table \ref{tab:main}, our \feedback{} feedback mechanism significantly enhances the factuality of summaries across all tested summarization LLMs on both datasets. For instance, within the SciMRC dataset, the implementation of \feedback{} feedback across three LLMs leads to an average factuality improvement of 6.6\% (54.02\% $\rightarrow$ 60.61\%) as measured by QAGS, and a 4.3\% increase (28.7\% $\rightarrow$ 33\%) as assessed by QuestEval. Likewise, the QASPER dataset demonstrates 5.3\% and 3.2\% improvement on QAGS and QuestEval, respectively. Furthermore, the average ROUGE score sees a modest enhancement of 0.97\%, suggesting that the text quality of summaries remains consistent, whether or not \feedback{} feedback is applied.




\section{Analysis}
\subsection{Comparing Feedback Mechanisms}
\begin{figure*}[h]
\centering
\includegraphics[width=0.79\textwidth,height=0.55\textwidth]{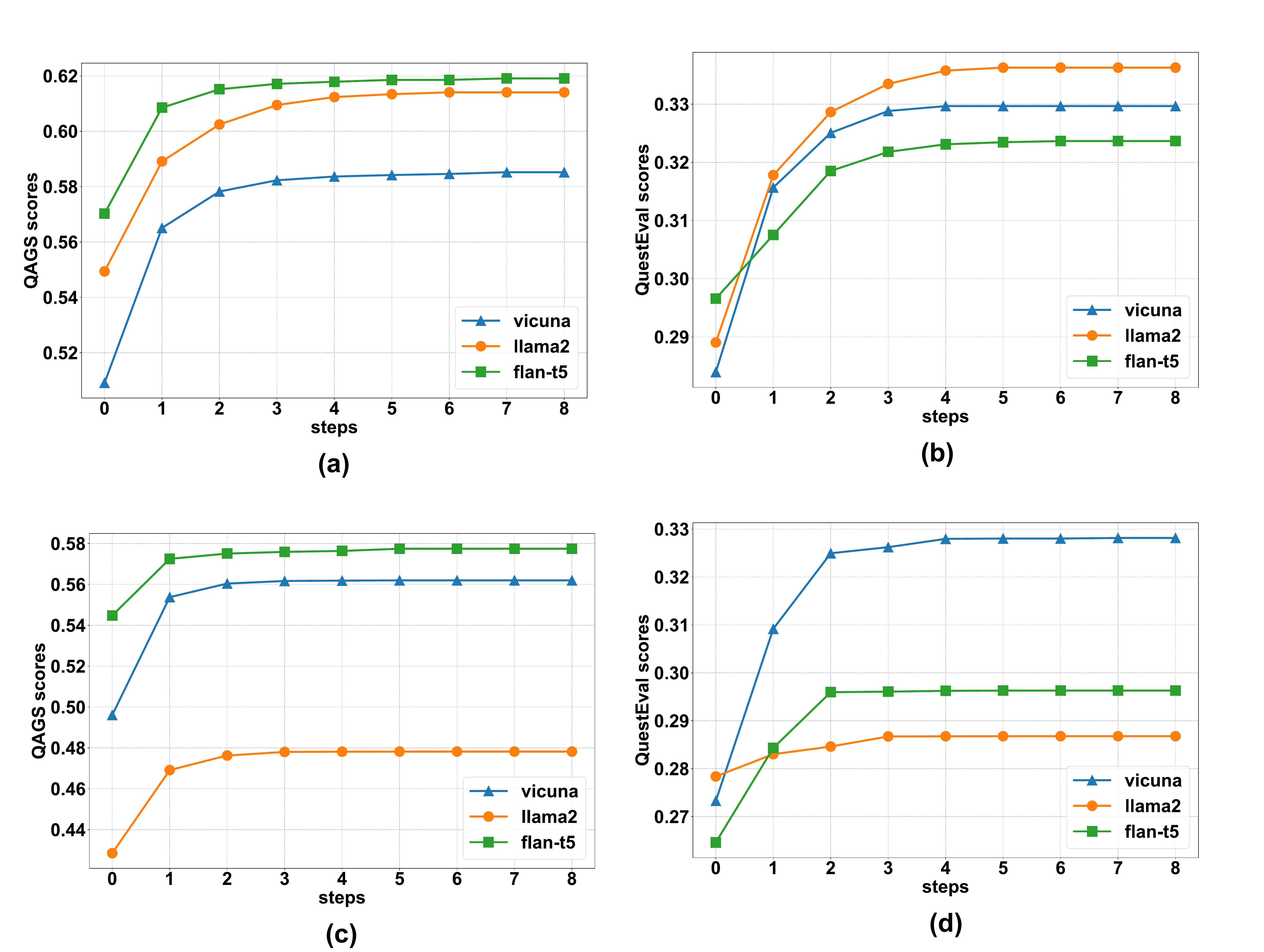}
\caption{Correlation between the factuality improvement and the number of refining steps on SciMRC. Figure (a) and (b) depict the QAGS score and QuestEval score on SciMRC. Figure (a) and (b) depict the QAGS score and QuestEval score on QASPER.}
\label{fig:multi}
\end{figure*}

To validate the efficacy of \feedback{} feedback, we employ two distinct methodologies: pre-hoc prompt engineering and post-hoc generic feedback. 
\textbf{Pre-hoc prompt engineering} is designed to preempt the generation of non-factual content by incorporating preventive measures into the initial prompt itself. Details on prompts used for prompt engineering are included in the Appendix~\ref{sec:ppe}.

In contrast, \textbf{Post-hoc generic feedback} refers to post-hoc feedback without detailed factual information and without providing actionable instructions. For the exact prompts we tested, please refer to Appendix~\ref{sec:fgp}.
These approaches enable us to compare the impact of \feedback{}'s tailored feedback against 
these alternative feedback methods.

Table~\ref{tab:fed} compares the performance of these two methods with \feedback{}. Note that we use the best QAGS and QuestEval scores among different pre-hoc prompts as the result of the Pre-hoc Prompt approach. While all forms provide some improvement over no feedback, we validate that post-hoc strategies indeed outperform the pre-hoc preventative strategy (compare Row~1 versus Rows~2~and~3).  Furthermore, detailed pinpoint feedback from \feedback{} yields larger gains still, as evidenced by the specific insights on content factuality, merely indicating factual inaccuracies and urging improvement.
This approach mirrors a teacher--student interaction, with \feedback{} acting as a guiding teacher, pinpointing not just whether the output is correct or not (as generic feedback does) but also the specifics of correctness and inaccuracy, which allow students to take appropriate actions.



\begin{table}[tb]
\centering
\small
\begin{tabular}{cccl}
\toprule
Feedback Type & QAGS & QuestEval \\ 
\midrule
Pre-hoc Prompt  & 57.68\%                  & 29.62\%   \\ 
Generic Feedback        & 59.88\%                  & 29.89\%   \\
\feedback{}                 & \textbf{61.91}\%                  & \textbf{32.42}\%   \\
\bottomrule
\end{tabular}
\caption{Results of different feedback strategies on SciMRC. The summarization model for demonstration is FLAN-T5.}
\label{tab:fed}
\end{table}

\subsection{Impact of Iterative Refining Process}
In this section, we analyze the impact of having multiple steps of refining and quantify factuality improvements along the iterative refining process. In Figure \ref{fig:multi}, we plot correlation curves between both QAGS and QuestEval scores and the number of refining steps. The model's proficiency in generating factually consistent summaries exhibits a noteworthy augmentation as the refinement progresses. This observed increase aligns with our expectations and underscores the model's adaptability to the evolving feedback content. 

Notably, around the fourth or fifth refinement step for all LLMs on both datasets, a discernible trend of convergence emerges in the factuality of the generated summaries. This temporal convergence implies that the summarization module has received feedback generated from all questions. Thus, there's no new feedback for further refinement.


\subsection{Different Feedback LLMs}
\label{sec:gpt}
In contrast to the reliance on large-scale commercial LLM products, a significant advantage of our method is the ability to achieve substantial improvements at a much lower cost by using a compact, open-source LLM for feedback. To demonstrate this, we replace the feedback model based on fine-tuned Vicuna with GPT-3.5. According to Table \ref{tab:gpt}, the factuality scores between the two models are closely matched, with GPT-3.5 leading by a slight 0.33\% on QuestEval and Vicuna outperforming by 0.68\% on QAGS. This suggests that the feedback quality from both systems is comparably effective. Our findings highlight that with proper designing, a fine-tuned open-source LLM with 7 billion parameters can match the performance of GPT-3.5\footnote{Although, we note that our feedback LLM may not compete with stronger products such as GPT-4.}. This not only proves to be cost-efficient but also ensures deterministic and reproducible outcomes.

\begin{table}[htb]
\centering
\small
\begin{tabular}{cccl}
\toprule
Feedback Model & QAGS & QuestEval        \\ \midrule
GPT-3.5                      & 61.41\%        & \textbf{32.70\%} \\
Vicuna                   & \textbf{61.93\%}         & 32.42\%          \\ 
\bottomrule
\end{tabular}
\caption{Results of comparison between different feedback models on SciMRC: fine-tuned Vicuna and GPT-3.5. The summarization model used is FLAN-T5.}
\label{tab:gpt}
\end{table}

\subsection{Importance of Question Quality}
In Section \ref{sec:fm}, we utilize scientific QA datasets for feedback generation, where the questions are human-annotated and considered to be of high quality. These datasets, including SciMRC and QASPER, have demonstrated their effectiveness in enabling feedback LLMs to provide informative factuality feedback to enhance summarization LLMs through previous experiments in Table~\ref{tab:main}. This section explores the potential of using system-generated questions. For this purpose, we generate questions for each scientific paper in the SciMRC dataset by extracting entities as answer candidates and then generating questions conditioned on these candidates. For simplicity, we employ the question-generation model proposed by~\citet{ushio-etal-2023-a-practical-toolkit}.

Upon replacing the original SciMRC dataset with the one generated by the question generation model for iterative factuality refining, the results---detailed in Table \ref{tab:qg}---show a significant decline in performance across both benchmarks when using model-generated questions, with QAGS falling even below the performance levels of models without any feedback. Thus, we argue that low-quality model-generated QA data is insufficient for high-quality feedback generation. Examples of both model-generated and human-annotated QA pairs are provided in the Appendix~\ref{ques_comp} for comparison.

\begin{table}[htb]
\centering
\small
\begin{tabular}{cccl}
\toprule
Source of QA       & QAGS & QuestEval        \\ 
\midrule
Model-generated & 56.64\%                  & 31.34\%         \\
Human-annotated & \textbf{61.91\%}         & \textbf{32.42\%} \\
\bottomrule
\end{tabular}
\caption{Comparison between different sources of QA data. Human-annotated data refers to data from SciMRC. The summarization model used is FLAN-T5.}
\label{tab:qg}
\end{table}

\begin{figure*}[t]
\centering
\includegraphics[width=1.0\textwidth,height=0.37\textwidth]{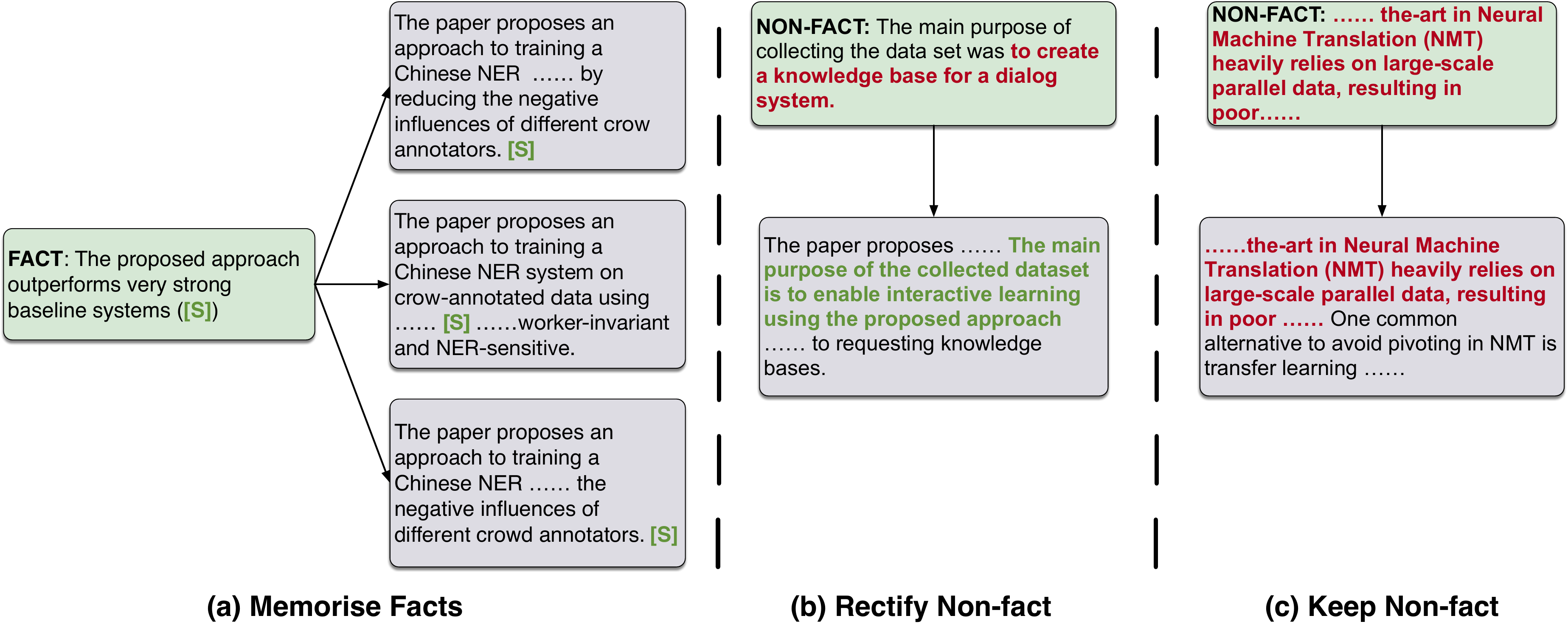}
\caption{Examples of how LLMs respond to {\color{ForestGreen} fact} and \textcolor{Maroon}{non-fact} feedback, respectively. In (a), \textbf{[S]} denotes the whole sentence. Arrows denote subsequent summaries after the model receives the feedback.}
\label{fig:case_study}
\end{figure*}

\subsection{Human Evaluation}
We conduct a human evaluation on a sampled subset (20 papers) of SciMRC to further validate the factuality enhancement by \feedback{}. Three qualified annotators are hired for the evaluation and paid at the local compensation level. Annotations on five papers are used to calculate inter-agreement between annotators, with the Cohen Kappa coefficient $\mathcal{K}$ being 0.51, indicating acceptable agreement. For each summary, its factuality score is computed as the ratio of factual sentences out of all. The factuality score of a system is calculated as the average factuality score of all summaries. According to the result, the proposed \feedback{} leads to an 11\% gain (79\% $\rightarrow$ 90\%) in the factuality over the baseline. We provide more details of human evaluation in the Appendix \ref{sec:human}.

\subsection{Case Study}
We provide a case study to demonstrate some insights into the behavior of LLMs during the iterative refining process. Since the feedback content is a combination of facts and non-facts found in the generated summaries, we show by case whether the model can understand the feedback and how the model makes decisions accordingly in Figure \ref{fig:case_study}. More cases are provided in Appendix \ref{sec:cases}.

\paragraph{LLM Memorizes Facts}
With regard to feedback pertaining to factual content, we observed that the LLM exhibits a tendency to memorize factual information. As depicted in Figure \ref{fig:case_study}(a), when the feedback highlights a factual signal such as ``The proposed approach outperforms very strong baseline systems,'' subsequent summaries generated by the summarization LLM, while not necessarily replicating the exact phrasing, consistently contain this factual statement. The decision made by the model to memorize facts, as provided by feedback, follows our act-on-feedback prompt and is deemed a logical strategy in the process of refining summary generation. This behavior not only underscores the model's capacity to respond to facts but also aligns with conventional human reasoning habits. 

\paragraph{LLM Struggles with Non-facts}
On the other hand, the performance of the LLM in handling non-facts leaves room for improvement, as it struggles to make nuanced judgments on how to address non-factual content provided in the feedback. Figure \ref{fig:case_study}(b) exemplifies a successful instance where the model accurately comprehends the instruction and rectifies the error. For example, when the feedback indicates that ``The main purpose of collecting the data set was to create a knowledge base for a dialog system.'' is a non-fact, LLMs are capable of executing the instruction to amend this non-fact as requested in the act-on-feedback prompt by identifying relevant content of the original paper then output the factual information.

However, in certain instances, the LLM faces challenges in making effective decisions regarding non-fact content. For instance, as depicted in Figure \ref{fig:case_study}(c), the summarization LLM retains the erroneous information in subsequent outputs despite the feedback explicitly stating the non-factual nature of a given statement. We attribute this phenomenon primarily to the LLM's extensive exposure to affirmative instructions during large-scale pre-training, such as ``answer this question,'' with comparatively limited exposure to negation or contradiction. This implies that the LLM may not excel in negating information, contributing to its occasional difficulty in appropriately handling non-facts provided in the feedback.

\section{Conclusion}
In this paper, we introduce a novel paradigm for leveraging LLM-generated \feedback{} feedback to enhance the factual consistency of scientific summarization. The method, \textit{Iterative Factuality Refinment on \feedback{} feedback} incorporates fine-grained factual feedback design and iterative refining process, enabling LLMs to directly learn from facts and non-facts and gradually improve the factual consistency. We show by extensive experiments that \feedback{} feedback can benefit various open-source summarization LLMs to refine their outputs. Furthermore, our approach shows the potential of compact LLMs to provide high-quality feedback that is comparable with GPT-3.5. Based on case studies, we propose future work on enabling LLMs to handle negative instructions. 

\section*{Acknowledgement}

This research is supported by the Singapore Ministry of Education Academic Research Fund Tier 1 (251RES2216). The computational work for this article was partially performed on resources of the National Supercomputing Centre, Singapore.
\section*{Limitations}
This study employs informative feedback through the fine-tuning of a compact LLM using high-quality, domain-specific question-answering data. However, the scarcity of such specialized datasets presents a significant obstacle. Consequently, the applicability of this method is domain-specific, contingent upon the availability of sufficient high-quality QA data for adaptation to other domains.

\bibliography{anthology,custom}

\begin{thebibliography}{38}
\expandafter\ifx\csname natexlab\endcsname\relax\def\natexlab#1{#1}\fi

\bibitem[{Aharoni et~al.(2023)Aharoni, Narayan, Maynez, Herzig, Clark, and Lapata}]{aharoni-etal-2023-multilingual}
Roee Aharoni, Shashi Narayan, Joshua Maynez, Jonathan Herzig, Elizabeth Clark, and Mirella Lapata. 2023.
\newblock \href {https://doi.org/10.18653/v1/2023.findings-acl.220} {Multilingual summarization with factual consistency evaluation}.
\newblock In \emph{Findings of the Association for Computational Linguistics: ACL 2023}, pages 3562--3591, Toronto, Canada. Association for Computational Linguistics.

\bibitem[{An et~al.(2023)An, Ma, Lin, Zheng, Lou, and Chen}]{learningfrommistake2023}
Shengnan An, Zexiong Ma, Zeqi Lin, Nanning Zheng, Jian{-}Guang Lou, and Weizhu Chen. 2023.
\newblock \href {https://doi.org/10.48550/ARXIV.2310.20689} {Learning from mistakes makes {LLM} better reasoner}.
\newblock \emph{CoRR}, abs/2310.20689.

\bibitem[{Balachandran et~al.(2022)Balachandran, Hajishirzi, Cohen, and Tsvetkov}]{balachandran-etal-2022-correcting}
Vidhisha Balachandran, Hannaneh Hajishirzi, William Cohen, and Yulia Tsvetkov. 2022.
\newblock \href {https://doi.org/10.18653/v1/2022.emnlp-main.667} {Correcting diverse factual errors in abstractive summarization via post-editing and language model infilling}.
\newblock In \emph{Proceedings of the 2022 Conference on Empirical Methods in Natural Language Processing}, pages 9818--9830, Abu Dhabi, United Arab Emirates. Association for Computational Linguistics.

\bibitem[{Bhattacharjee et~al.(2023)Bhattacharjee, Hasan, Ahmad, Li, Kang, and Shahriyar}]{bhattacharjee-etal-2023-crosssum}
Abhik Bhattacharjee, Tahmid Hasan, Wasi~Uddin Ahmad, Yuan-Fang Li, Yong-Bin Kang, and Rifat Shahriyar. 2023.
\newblock \href {https://doi.org/10.18653/v1/2023.acl-long.143} {{C}ross{S}um: Beyond {E}nglish-centric cross-lingual summarization for 1,500+ language pairs}.
\newblock In \emph{Proceedings of the 61st Annual Meeting of the Association for Computational Linguistics (Volume 1: Long Papers)}, pages 2541--2564, Toronto, Canada. Association for Computational Linguistics.

\bibitem[{Bird et~al.(2009)Bird, Klein, and Loper}]{bird2009natural}
Steven Bird, Ewan Klein, and Edward Loper. 2009.
\newblock \emph{Natural language processing with Python: analyzing text with the natural language toolkit}.
\newblock " O'Reilly Media, Inc.".

\bibitem[{Chen et~al.(2023{\natexlab{a}})Chen, Guo, Haddow, and Heafield}]{chen2023iterative}
Pinzhen Chen, Zhicheng Guo, Barry Haddow, and Kenneth Heafield. 2023{\natexlab{a}}.
\newblock \href {http://arxiv.org/abs/2306.03856} {Iterative translation refinement with large language models}.

\bibitem[{Chen et~al.(2021)Chen, Liu, and Qiu}]{chen-etal-2021-factuality-checkers}
Yiran Chen, Pengfei Liu, and Xipeng Qiu. 2021.
\newblock \href {https://doi.org/10.18653/v1/2021.findings-emnlp.179} {Are factuality checkers reliable? adversarial meta-evaluation of factuality in summarization}.
\newblock In \emph{Findings of the Association for Computational Linguistics: EMNLP 2021}, pages 2082--2095, Punta Cana, Dominican Republic. Association for Computational Linguistics.

\bibitem[{Chen et~al.(2023{\natexlab{b}})Chen, Liu, Xu, Yang, Zhu, Zeng, and Zhang}]{chen-etal-2023-unisumm}
Yulong Chen, Yang Liu, Ruochen Xu, Ziyi Yang, Chenguang Zhu, Michael Zeng, and Yue Zhang. 2023{\natexlab{b}}.
\newblock \href {https://doi.org/10.18653/v1/2023.acl-long.718} {{U}ni{S}umm and {S}umm{Z}oo: Unified model and diverse benchmark for few-shot summarization}.
\newblock In \emph{Proceedings of the 61st Annual Meeting of the Association for Computational Linguistics (Volume 1: Long Papers)}, pages 12833--12855, Toronto, Canada. Association for Computational Linguistics.

\bibitem[{Chiang et~al.(2023)Chiang, Li, Lin, Sheng, Wu, Zhang, Zheng, Zhuang, Zhuang, Gonzalez, Stoica, and Xing}]{vicuna2023}
Wei-Lin Chiang, Zhuohan Li, Zi~Lin, Ying Sheng, Zhanghao Wu, Hao Zhang, Lianmin Zheng, Siyuan Zhuang, Yonghao Zhuang, Joseph~E. Gonzalez, Ion Stoica, and Eric~P. Xing. 2023.
\newblock \href {https://lmsys.org/blog/2023-03-30-vicuna/} {Vicuna: An open-source chatbot impressing gpt-4 with 90\%* chatgpt quality}.

\bibitem[{Chung et~al.(2022)Chung, Hou, Longpre, Zoph, Tay, Fedus, Li, Wang, Dehghani, Brahma, Webson, Gu, Dai, Suzgun, Chen, Chowdhery, Castro-Ros, Pellat, Robinson, Valter, Narang, Mishra, Yu, Zhao, Huang, Dai, Yu, Petrov, Chi, Dean, Devlin, Roberts, Zhou, Le, and Wei}]{chung2022scaling}
Hyung~Won Chung, Le~Hou, Shayne Longpre, Barret Zoph, Yi~Tay, William Fedus, Yunxuan Li, Xuezhi Wang, Mostafa Dehghani, Siddhartha Brahma, Albert Webson, Shixiang~Shane Gu, Zhuyun Dai, Mirac Suzgun, Xinyun Chen, Aakanksha Chowdhery, Alex Castro-Ros, Marie Pellat, Kevin Robinson, Dasha Valter, Sharan Narang, Gaurav Mishra, Adams Yu, Vincent Zhao, Yanping Huang, Andrew Dai, Hongkun Yu, Slav Petrov, Ed~H. Chi, Jeff Dean, Jacob Devlin, Adam Roberts, Denny Zhou, Quoc~V. Le, and Jason Wei. 2022.
\newblock \href {http://arxiv.org/abs/2210.11416} {Scaling instruction-finetuned language models}.

\bibitem[{Dasigi et~al.(2021)Dasigi, Lo, Beltagy, Cohan, Smith, and Gardner}]{dasigi-etal-2021-dataset}
Pradeep Dasigi, Kyle Lo, Iz~Beltagy, Arman Cohan, Noah~A. Smith, and Matt Gardner. 2021.
\newblock \href {https://doi.org/10.18653/v1/2021.naacl-main.365} {A dataset of information-seeking questions and answers anchored in research papers}.
\newblock In \emph{Proceedings of the 2021 Conference of the North American Chapter of the Association for Computational Linguistics: Human Language Technologies}, pages 4599--4610, Online. Association for Computational Linguistics.

\bibitem[{Du et~al.(2023)Du, Li, Torralba, Tenenbaum, and Mordatch}]{du2023improving}
Yilun Du, Shuang Li, Antonio Torralba, Joshua~B Tenenbaum, and Igor Mordatch. 2023.
\newblock Improving factuality and reasoning in language models through multiagent debate.
\newblock \emph{arXiv preprint arXiv:2305.14325}.

\bibitem[{Gou et~al.(2024)Gou, Shao, Gong, yelong shen, Yang, Duan, and Chen}]{gou2024critic}
Zhibin Gou, Zhihong Shao, Yeyun Gong, yelong shen, Yujiu Yang, Nan Duan, and Weizhu Chen. 2024.
\newblock \href {https://openreview.net/forum?id=Sx038qxjek} {{CRITIC}: Large language models can self-correct with tool-interactive critiquing}.
\newblock In \emph{The Twelfth International Conference on Learning Representations}.

\bibitem[{He et~al.(2023)He, Peng, Wang, Liu, Xu, Hassan, Shi, Zhu, Xiong, Zeng, Gao, and Huang}]{he-etal-2023-z}
Pengcheng He, Baolin Peng, Song Wang, Yang Liu, Ruochen Xu, Hany Hassan, Yu~Shi, Chenguang Zhu, Wayne Xiong, Michael Zeng, Jianfeng Gao, and Xuedong Huang. 2023.
\newblock \href {https://doi.org/10.18653/v1/2023.acl-long.279} {{Z}-code++: A pre-trained language model optimized for abstractive summarization}.
\newblock In \emph{Proceedings of the 61st Annual Meeting of the Association for Computational Linguistics (Volume 1: Long Papers)}, pages 5095--5112, Toronto, Canada. Association for Computational Linguistics.

\bibitem[{Honnibal and Montani(2017)}]{spacy2}
Matthew Honnibal and Ines Montani. 2017.
\newblock {spaCy 2}: Natural language understanding with {B}loom embeddings, convolutional neural networks and incremental parsing.
\newblock To appear.

\bibitem[{Hu et~al.(2022)Hu, yelong shen, Wallis, Allen-Zhu, Li, Wang, Wang, and Chen}]{hu2022lora}
Edward~J Hu, yelong shen, Phillip Wallis, Zeyuan Allen-Zhu, Yuanzhi Li, Shean Wang, Lu~Wang, and Weizhu Chen. 2022.
\newblock \href {https://openreview.net/forum?id=nZeVKeeFYf9} {Lo{RA}: Low-rank adaptation of large language models}.
\newblock In \emph{International Conference on Learning Representations}.

\bibitem[{Huang et~al.(2022)Huang, Gu, Hou, Wu, Wang, Yu, and Han}]{huang2022large}
Jiaxin Huang, Shixiang~Shane Gu, Le~Hou, Yuexin Wu, Xuezhi Wang, Hongkun Yu, and Jiawei Han. 2022.
\newblock Large language models can self-improve.
\newblock \emph{arXiv preprint arXiv:2210.11610}.

\bibitem[{Krishna et~al.(2023)Krishna, Bransom, Kuehl, Iyyer, Dasigi, Cohan, and Lo}]{krishna-etal-2023-longeval}
Kalpesh Krishna, Erin Bransom, Bailey Kuehl, Mohit Iyyer, Pradeep Dasigi, Arman Cohan, and Kyle Lo. 2023.
\newblock \href {https://doi.org/10.18653/v1/2023.eacl-main.121} {{L}ong{E}val: Guidelines for human evaluation of faithfulness in long-form summarization}.
\newblock In \emph{Proceedings of the 17th Conference of the European Chapter of the Association for Computational Linguistics}, pages 1650--1669, Dubrovnik, Croatia. Association for Computational Linguistics.

\bibitem[{Lin(2004)}]{lin-2004-ROUGE}
Chin-Yew Lin. 2004.
\newblock \href {https://aclanthology.org/W04-1013} {{ROUGE}: A package for automatic evaluation of summaries}.
\newblock In \emph{Text Summarization Branches Out}, pages 74--81, Barcelona, Spain. Association for Computational Linguistics.

\bibitem[{Lin et~al.(2022)Lin, Hilton, and Evans}]{lin-etal-2022-truthfulqa}
Stephanie Lin, Jacob Hilton, and Owain Evans. 2022.
\newblock \href {https://doi.org/10.18653/v1/2022.acl-long.229} {{T}ruthful{QA}: Measuring how models mimic human falsehoods}.
\newblock In \emph{Proceedings of the 60th Annual Meeting of the Association for Computational Linguistics (Volume 1: Long Papers)}, pages 3214--3252, Dublin, Ireland. Association for Computational Linguistics.

\bibitem[{Liu et~al.(2023)Liu, Deb, Teruel, Halfaker, Radev, and Awadallah}]{liu-etal-2023-improving}
Yixin Liu, Budhaditya Deb, Milagro Teruel, Aaron Halfaker, Dragomir Radev, and Ahmed~Hassan Awadallah. 2023.
\newblock \href {https://doi.org/10.18653/v1/2023.acl-long.844} {On improving summarization factual consistency from natural language feedback}.
\newblock In \emph{Proceedings of the 61st Annual Meeting of the Association for Computational Linguistics (Volume 1: Long Papers)}, pages 15144--15161, Toronto, Canada. Association for Computational Linguistics.

\bibitem[{Madaan et~al.(2023)Madaan, Tandon, Gupta, Hallinan, Gao, Wiegreffe, Alon, Dziri, Prabhumoye, Yang, Gupta, Majumder, Hermann, Welleck, Yazdanbakhsh, and Clark}]{madaan2023selfrefine}
Aman Madaan, Niket Tandon, Prakhar Gupta, Skyler Hallinan, Luyu Gao, Sarah Wiegreffe, Uri Alon, Nouha Dziri, Shrimai Prabhumoye, Yiming Yang, Shashank Gupta, Bodhisattwa~Prasad Majumder, Katherine Hermann, Sean Welleck, Amir Yazdanbakhsh, and Peter Clark. 2023.
\newblock \href {https://openreview.net/forum?id=S37hOerQLB} {Self-refine: Iterative refinement with self-feedback}.
\newblock In \emph{Thirty-seventh Conference on Neural Information Processing Systems}.

\bibitem[{Min et~al.(2023)Min, Krishna, Lyu, Lewis, Yih, Koh, Iyyer, Zettlemoyer, and Hajishirzi}]{min-etal-2023-factscore}
Sewon Min, Kalpesh Krishna, Xinxi Lyu, Mike Lewis, Wen-tau Yih, Pang Koh, Mohit Iyyer, Luke Zettlemoyer, and Hannaneh Hajishirzi. 2023.
\newblock \href {https://doi.org/10.18653/v1/2023.emnlp-main.741} {{FA}ct{S}core: Fine-grained atomic evaluation of factual precision in long form text generation}.
\newblock In \emph{Proceedings of the 2023 Conference on Empirical Methods in Natural Language Processing}, pages 12076--12100, Singapore. Association for Computational Linguistics.

\bibitem[{Nathani et~al.(2023)Nathani, Wang, Pan, and Wang}]{nathani-etal-2023-maf}
Deepak Nathani, David Wang, Liangming Pan, and William Wang. 2023.
\newblock \href {https://doi.org/10.18653/v1/2023.emnlp-main.407} {{MAF}: Multi-aspect feedback for improving reasoning in large language models}.
\newblock In \emph{Proceedings of the 2023 Conference on Empirical Methods in Natural Language Processing}, pages 6591--6616, Singapore. Association for Computational Linguistics.

\bibitem[{Pan et~al.(2023)Pan, Saxon, Xu, Nathani, Wang, and Wang}]{pan2023automatically}
Liangming Pan, Michael Saxon, Wenda Xu, Deepak Nathani, Xinyi Wang, and William~Yang Wang. 2023.
\newblock Automatically correcting large language models: Surveying the landscape of diverse self-correction strategies.
\newblock \emph{arXiv preprint arXiv:2308.03188}.

\bibitem[{Roit et~al.(2023)Roit, Ferret, Shani, Aharoni, Cideron, Dadashi, Geist, Girgin, Hussenot, Keller, Momchev, Ramos~Garea, Stanczyk, Vieillard, Bachem, Elidan, Hassidim, Pietquin, and Szpektor}]{roit-etal-2023-factually}
Paul Roit, Johan Ferret, Lior Shani, Roee Aharoni, Geoffrey Cideron, Robert Dadashi, Matthieu Geist, Sertan Girgin, Leonard Hussenot, Orgad Keller, Nikola Momchev, Sabela Ramos~Garea, Piotr Stanczyk, Nino Vieillard, Olivier Bachem, Gal Elidan, Avinatan Hassidim, Olivier Pietquin, and Idan Szpektor. 2023.
\newblock \href {https://doi.org/10.18653/v1/2023.acl-long.344} {Factually consistent summarization via reinforcement learning with textual entailment feedback}.
\newblock In \emph{Proceedings of the 61st Annual Meeting of the Association for Computational Linguistics (Volume 1: Long Papers)}, pages 6252--6272, Toronto, Canada. Association for Computational Linguistics.

\bibitem[{Scialom et~al.(2021)Scialom, Dray, Lamprier, Piwowarski, Staiano, Wang, and Gallinari}]{scialom-etal-2021-questeval}
Thomas Scialom, Paul-Alexis Dray, Sylvain Lamprier, Benjamin Piwowarski, Jacopo Staiano, Alex Wang, and Patrick Gallinari. 2021.
\newblock \href {https://doi.org/10.18653/v1/2021.emnlp-main.529} {{Q}uest{E}val: Summarization asks for fact-based evaluation}.
\newblock In \emph{Proceedings of the 2021 Conference on Empirical Methods in Natural Language Processing}, pages 6594--6604, Online and Punta Cana, Dominican Republic. Association for Computational Linguistics.

\bibitem[{Scialom et~al.(2019)Scialom, Lamprier, Piwowarski, and Staiano}]{scialom-etal-2019-answers}
Thomas Scialom, Sylvain Lamprier, Benjamin Piwowarski, and Jacopo Staiano. 2019.
\newblock \href {https://doi.org/10.18653/v1/D19-1320} {Answers unite! unsupervised metrics for reinforced summarization models}.
\newblock In \emph{Proceedings of the 2019 Conference on Empirical Methods in Natural Language Processing and the 9th International Joint Conference on Natural Language Processing (EMNLP-IJCNLP)}, pages 3246--3256, Hong Kong, China. Association for Computational Linguistics.

\bibitem[{Tang et~al.(2023)Tang, Goyal, Fabbri, Laban, Xu, Yavuz, Kryscinski, Rousseau, and Durrett}]{tang-etal-2023-understanding}
Liyan Tang, Tanya Goyal, Alex Fabbri, Philippe Laban, Jiacheng Xu, Semih Yavuz, Wojciech Kryscinski, Justin Rousseau, and Greg Durrett. 2023.
\newblock \href {https://doi.org/10.18653/v1/2023.acl-long.650} {Understanding factual errors in summarization: Errors, summarizers, datasets, error detectors}.
\newblock In \emph{Proceedings of the 61st Annual Meeting of the Association for Computational Linguistics (Volume 1: Long Papers)}, pages 11626--11644, Toronto, Canada. Association for Computational Linguistics.

\bibitem[{Touvron et~al.(2023)Touvron, Martin, Stone, Albert, Almahairi, Babaei, Bashlykov, Batra, Bhargava, Bhosale, Bikel, Blecher, Ferrer, Chen, Cucurull, Esiobu, Fernandes, Fu, Fu, Fuller, Gao, Goswami, Goyal, Hartshorn, Hosseini, Hou, Inan, Kardas, Kerkez, Khabsa, Kloumann, Korenev, Koura, Lachaux, Lavril, Lee, Liskovich, Lu, Mao, Martinet, Mihaylov, Mishra, Molybog, Nie, Poulton, Reizenstein, Rungta, Saladi, Schelten, Silva, Smith, Subramanian, Tan, Tang, Taylor, Williams, Kuan, Xu, Yan, Zarov, Zhang, Fan, Kambadur, Narang, Rodriguez, Stojnic, Edunov, and Scialom}]{touvron2023llama}
Hugo Touvron, Louis Martin, Kevin Stone, Peter Albert, Amjad Almahairi, Yasmine Babaei, Nikolay Bashlykov, Soumya Batra, Prajjwal Bhargava, Shruti Bhosale, Dan Bikel, Lukas Blecher, Cristian~Canton Ferrer, Moya Chen, Guillem Cucurull, David Esiobu, Jude Fernandes, Jeremy Fu, Wenyin Fu, Brian Fuller, Cynthia Gao, Vedanuj Goswami, Naman Goyal, Anthony Hartshorn, Saghar Hosseini, Rui Hou, Hakan Inan, Marcin Kardas, Viktor Kerkez, Madian Khabsa, Isabel Kloumann, Artem Korenev, Punit~Singh Koura, Marie-Anne Lachaux, Thibaut Lavril, Jenya Lee, Diana Liskovich, Yinghai Lu, Yuning Mao, Xavier Martinet, Todor Mihaylov, Pushkar Mishra, Igor Molybog, Yixin Nie, Andrew Poulton, Jeremy Reizenstein, Rashi Rungta, Kalyan Saladi, Alan Schelten, Ruan Silva, Eric~Michael Smith, Ranjan Subramanian, Xiaoqing~Ellen Tan, Binh Tang, Ross Taylor, Adina Williams, Jian~Xiang Kuan, Puxin Xu, Zheng Yan, Iliyan Zarov, Yuchen Zhang, Angela Fan, Melanie Kambadur, Sharan Narang, Aurelien Rodriguez, Robert Stojnic, Sergey Edunov, and Thomas
  Scialom. 2023.
\newblock \href {http://arxiv.org/abs/2307.09288} {Llama 2: Open foundation and fine-tuned chat models}.

\bibitem[{Ushio et~al.(2023)Ushio, Alva-Manchego, and Camacho-Collados}]{ushio-etal-2023-a-practical-toolkit}
Asahi Ushio, Fernando Alva-Manchego, and Jose Camacho-Collados. 2023.
\newblock A practical toolkit for multilingual question and answer generation.
\newblock In \emph{Proceedings of the 61th Annual Meeting of the Association for Computational Linguistics: System Demonstrations}, Toronto, Canada. Association for Computational Linguistics.

\bibitem[{Vaswani et~al.(2017)Vaswani, Shazeer, Parmar, Uszkoreit, Jones, Gomez, Kaiser, and Polosukhin}]{NIPS2017_3f5ee243}
Ashish Vaswani, Noam Shazeer, Niki Parmar, Jakob Uszkoreit, Llion Jones, Aidan~N Gomez, \L~ukasz Kaiser, and Illia Polosukhin. 2017.
\newblock \href {https://proceedings.neurips.cc/paper_files/paper/2017/file/3f5ee243547dee91fbd053c1c4a845aa-Paper.pdf} {Attention is all you need}.
\newblock In \emph{Advances in Neural Information Processing Systems}, volume~30. Curran Associates, Inc.

\bibitem[{Wang et~al.(2020)Wang, Cho, and Lewis}]{wang2020asking}
Alex Wang, Kyunghyun Cho, and Mike Lewis. 2020.
\newblock Asking and answering questions to evaluate the factual consistency of summaries.
\newblock \emph{arXiv preprint arXiv:2004.04228}.

\bibitem[{Welleck et~al.(2022)Welleck, Lu, West, Brahman, Shen, Khashabi, and Choi}]{welleck2022generating}
Sean Welleck, Ximing Lu, Peter West, Faeze Brahman, Tianxiao Shen, Daniel Khashabi, and Yejin Choi. 2022.
\newblock Generating sequences by learning to self-correct.
\newblock \emph{arXiv preprint arXiv:2211.00053}.

\bibitem[{Wolf et~al.(2020)Wolf, Debut, Sanh, Chaumond, Delangue, Moi, Cistac, Rault, Louf, Funtowicz, Davison, Shleifer, von Platen, Ma, Jernite, Plu, Xu, Le~Scao, Gugger, Drame, Lhoest, and Rush}]{wolf-etal-2020-transformers}
Thomas Wolf, Lysandre Debut, Victor Sanh, Julien Chaumond, Clement Delangue, Anthony Moi, Pierric Cistac, Tim Rault, Remi Louf, Morgan Funtowicz, Joe Davison, Sam Shleifer, Patrick von Platen, Clara Ma, Yacine Jernite, Julien Plu, Canwen Xu, Teven Le~Scao, Sylvain Gugger, Mariama Drame, Quentin Lhoest, and Alexander Rush. 2020.
\newblock \href {https://doi.org/10.18653/v1/2020.emnlp-demos.6} {Transformers: State-of-the-art natural language processing}.
\newblock In \emph{Proceedings of the 2020 Conference on Empirical Methods in Natural Language Processing: System Demonstrations}, pages 38--45, Online. Association for Computational Linguistics.

\bibitem[{Yang et~al.(2018)Yang, Qi, Zhang, Bengio, Cohen, Salakhutdinov, and Manning}]{yang-etal-2018-hotpotqa}
Zhilin Yang, Peng Qi, Saizheng Zhang, Yoshua Bengio, William Cohen, Ruslan Salakhutdinov, and Christopher~D. Manning. 2018.
\newblock \href {https://doi.org/10.18653/v1/D18-1259} {{H}otpot{QA}: A dataset for diverse, explainable multi-hop question answering}.
\newblock In \emph{Proceedings of the 2018 Conference on Empirical Methods in Natural Language Processing}, pages 2369--2380, Brussels, Belgium. Association for Computational Linguistics.

\bibitem[{Zhang et~al.(2023{\natexlab{a}})Zhang, Press, Merrill, Liu, and Smith}]{zhang2023language}
Muru Zhang, Ofir Press, William Merrill, Alisa Liu, and Noah~A. Smith. 2023{\natexlab{a}}.
\newblock \href {http://arxiv.org/abs/2305.13534} {How language model hallucinations can snowball}.

\bibitem[{Zhang et~al.(2023{\natexlab{b}})Zhang, Zheng, Nie, Huang, and Mao}]{zhang2023scimrc}
Xiao Zhang, Heqi Zheng, Yuxiang Nie, Heyan Huang, and Xian-Ling Mao. 2023{\natexlab{b}}.
\newblock Scimrc: Multi-perspective scientific machine reading comprehension.
\newblock \emph{arXiv preprint arXiv:2306.14149}.

\end{thebibliography}

\appendix

\section*{Appendices}

\section{Human Evaluation}
\label{sec:human}
 
\paragraph{Annotator Selection} Annotators selected are required to have decent English reading ability, and all have obtained a Bachelor's degree. To increase the diversity, we hired annotators with different majors, including Computer Science and Material Science. Before completing the full evaluation, each annotator was asked to finish a training process by evaluating one paper. 

\paragraph{Annotation Data} The annotation data are summaries generated by Llama2 with and without \feedback{} feedback. On average, there are 5 sentences in each summary. The average length of the source text is 6000 characters.

\paragraph{Compensation} The rate of compensation is 15 SGD per hour, and we compensate annotators based on an average evaluation time.

\paragraph{Ethical Consideration}
In this study, strict ethical guidelines were adhered with the research protocol reviewed and approved by the Institutional Review Board (IRB) in our organization.
For example, informed consent was obtained from all participants, who were adequately informed about the study's purpose, procedures, potential risks, and benefits, as well as their right to withdraw at any time without any consequences. 



\section{Main Experiment Settings}
Experiments can be done on 2x A100 GPUs with 70 GPU hours in total. We list parameters used for experiments in Table \ref{tab:param}. We used NLTK~\cite{bird2009natural} and spaCy~\cite{spacy2} toolkits for text processing. Code is implemented in PyTorch~\cite{NIPS2017_3f5ee243} and Huggingface's Transformers~\cite{wolf-etal-2020-transformers}.

\begin{table}[htb]
\centering
\begin{tabular}{lr}
\toprule
Parameters  & Value \\ \midrule
num beams                                                       & 2     \\
temperarture                                                    & 1.3   \\
no repeated n-gram                                              & 3     \\ 
\hline
\multicolumn{2}{c}{\textit{Summarization LLM}} \\
\hline
max input length & 2048  \\
max new tokens   & 200   \\
min new tokens   & 100   \\
\hline
\multicolumn{2}{c}{\textit{Feedback LLM}} \\
\hline
max input length & 512   \\
max new tokens   & 100   \\
min new tokens   & 10    \\
\hline
\multicolumn{2}{c}{\textit{Iterative Factuality Refining}} \\
\hline
\# refining steps & 8 \\
batch size       & 4 \\
\bottomrule
\end{tabular}
\caption{Experimental parameters.}
\label{tab:param}
\end{table}

\section{Feedback LLM Fine-tuning}
\label{sec:fbft}
Below, we illustrate details of feedback LLM fine-tuning.
\begin{enumerate}
    \item \textbf{Fine-tuning objective}: The aim of fine-tuning the feedback LLM is to adapt it to the new task format we define, which we found open-source LLMs such as Vicuna-7b do not perform satisfactorily on. As described by Equation~\ref{eq:fb}, given a question paired with context, the model is expected to predict both the answer and the evidence. 

    \item \textbf{Dataset}:  We use HotpotQA~\cite{yang-etal-2018-hotpotqa} as the dataset for fine-tuning. As depicted in Table~\ref{tab:ftex}, inputs are context and questions and labels are answers and supporting facts. 

    \item \textbf{Fine-tune technique}:  LoRA~\citep{hu2022lora} is employed for fine-tuning the feedback LLM.

    \item \textbf{Hyperparameters}: See Table~\ref{tab:fth} for hyperparameter used during fine-tuning.

    \begin{table}[htb]
        \centering
        \begin{tabular}{lll}
            \hline
            \multicolumn{2}{c}{Parameters}       & Value \\ \hline
            \multicolumn{2}{c}{Max input length} & 512   \\
            \multicolumn{2}{l}{Max new tokens}   & 100   \\
            \multicolumn{2}{l}{Min new tokens}   & 10    \\
            \multicolumn{2}{l}{Learning rate}    & 3e-4  \\
            \multicolumn{2}{l}{Lora r}           & 8     \\
            \multicolumn{2}{l}{Lora alpha}       & 16    \\
            \multicolumn{2}{l}{Epochs}           & 10    \\ \hline
        \end{tabular}
        \caption{Hyperparameters for feedback LLM fine-tuning.}
        \label{tab:fth}
    \end{table}
    
    \item \textbf{Fine-tune effect}: In Table~\ref{tab:ftp}, we show the comparison between vanilla Vicuna's performance on SciMRC and fine-tuned Vicuna's performance on SciMRC. Here, we focus on consistency, which refers to the ratio of outputs that match our required output format in the prompt. We observe the vanilla LLM fails to adhere to the given output format by generating either only answers or evidence or indistinguishable answers and evidence. While the fine-tuned feedback model can provide outputs consistent with our required format.
\end{enumerate}

\begin{table*}[]
\centering
\small
\begin{tabular}{cl|cl}
\hline
\multicolumn{2}{c|}{Input}                                                                                                                                                                                                                                                                                                             & \multicolumn{2}{c}{Output}                                                                                                                                                                                     \\ \hline
Question                                                                                                                     & \multicolumn{1}{c|}{Context}                                                                                                                                                                            & Answer                                                                                     & \multicolumn{1}{c}{Supporting Fact}                                                                                      \\ \hline
\multicolumn{1}{l}{\begin{tabular}[c]{@{}l@{}}Who was born first, \\ Arthur Conan Doyle \\ or Penelope Lively?\end{tabular}} & \begin{tabular}[c]{@{}l@{}}Sir Arthur Ignatius \\ Conan Doyle KStJ, \\ DL was a British \\ writer best known for \\ his detective fiction \\ featuring the character \\ Sherlock Holmes...\end{tabular} & \multicolumn{1}{l}{\begin{tabular}[c]{@{}l@{}}Arthur Ignatius \\ Conan Doyle\end{tabular}} & \begin{tabular}[c]{@{}l@{}}Sir Arthur Ignatius \\ Conan Doyle KStJ, \\ DL was a British \\ writer...\end{tabular} \\ \hline
\end{tabular}
\caption{An example data point for feedback LLM fine-tuning. Input includes Question and Context. The model predicts the Answer with Evidence sentence accordingly.}
\label{tab:ftex}
\end{table*}

\begin{table*}[t]
\centering
\small
\resizebox{0.6\textwidth}{!}{%
\begin{tabular}{cccc}
\hline
\multicolumn{1}{l}{Feedback LLM} & \begin{tabular}[c]{@{}c@{}}Answer \\ ROUGE\end{tabular} & \begin{tabular}[c]{@{}c@{}}Evidence \\ ROUGE\end{tabular} & \multicolumn{1}{l}{Consistency} \\ \hline
\textit{Vanilla}                 & 28.82\%                                                 & 21.24\%                                                   & 47.1\%                          \\
\textit{Fine-tuned}              & \textbf{32.83\%}                                        & \textbf{36.41\%}                                          & \textbf{76.8\%}                 \\ \hline
\end{tabular}
}

\caption{Performance comparison on SciMRC between vanilla feedback LLM and fine-tuned feedback LLM. "Consistency" refers to the percentage of output that matches the expected output format in which answers and evidence can be decoded uniformly.}
\label{tab:ftp}
\end{table*}

\section{Question Comparison}
\label{ques_comp}
Table \ref{tab:qg_example} provides a comparison between human-annotated questions from the SciMRC dataset and questions generated by language models. We randomly pick three papers for demonstration and select 5 questions of each category. In general, human-annotated questions are more difficult and require a deeper understanding of the context. For example, to answer the fifth dataset question of paper A, the model needs to conduct multi-hop reasoning: Tt first locates two paper snippets for the authors' proposed method and traditional transfer methods, respectively. Then, it compares two pieces of information to derive the differences between them. However, model-generated questions don't require such complex reasoning chains.

\begin{table*}[ht]
\begin{tabular}{c|l|ll}
\hline
\multicolumn{1}{l|}{Papers} & \multicolumn{1}{c|}{Human-annotated Questions}                                                                                                                                                                                                                                                                                                                                                                                                                             & \multicolumn{2}{c}{Model-generated Questions}                                                                                                                                                                                                                                                                                                                                                                                                                                      \\ \hline
A                           & \begin{tabular}[c]{@{}l@{}}1. What do the authors argue about \\ the reason for zero-shot translation \\ failure in transfer learning?\\ \\ 2. In what scenario, does transfer \\ learning being explored?\\ \\ 3. What new transfer learning approach \\ do authors propose for NMT?\\ \\ 4. What novel pre-training method do \\ the authors propose?\\ \\ 5. How does the authors' proposed method \\ different from traditional transfer methods?\end{tabular} & \multicolumn{2}{l}{\begin{tabular}[c]{@{}l@{}}1. Neural Machine Translation has \\ dominated recent research on what?\\ \\ 2. Who proposes a new transfer learning \\ approach for NMT?\\ \\ 3. What does our proposed method \\ belong to?\\ \\ 4. What hinders transfer learning for \\ NMT towards the zero-resource setting?\\ \\ 5. Which model benefits from the \\ availability of the universal encoder?\end{tabular}}                                               \\ \hline
B                           & \begin{tabular}[c]{@{}l@{}}1. What corpus is the model based on?\\ \\ 2. What about the inter-annotator \\ agreement for the human evaluation?\\ \\ 3. What is the size of the corpus for \\ training?\\ \\ 4. Is the model able to predict dogmatism \\ in comments?\\ \\ 5. How are annotation results guaranteed?\end{tabular}                                                                                                                                  & \multicolumn{2}{l}{\begin{tabular}[c]{@{}l@{}}1. What does dogmatism influence \\ on social media?\\ \\ 2. What did MS produce?\\ \\ 3. How many Reddit posts are used \\ to train our model?\\ \\ 4. What does a comment not convey?\\ \\ 5. What number indicates agreement \\ indistinguishable from chance?\end{tabular}}                                                                                                                                                \\ \hline
C                           & \begin{tabular}[c]{@{}l@{}}1. How to improve the performance on \\ the child task?\\ \\ 2. Is it common to have source and \\ assisting languages with different word \\ order in a practical setting?\\ \\ 3. How to address the word order \\ divergence in this paper?\\ \\ 4. How to overcome data sparsity between \\ the source and the target languages?\\ \\ 5. What is the assisting language in all \\ experiments?\end{tabular}                         & \multicolumn{2}{l}{\begin{tabular}[c]{@{}l@{}}1. What is the de facto approach for \\ any NLP task?\\ \\ 2. What is the word order of the source \\ language?\\ \\ 3. Which BIBREF takes advantage of \\ lexical similarity between languages?\\ \\ 4. What did BIBREF10 propose to use \\ to represent the sentences in both the \\ parent and the child language?\\ \\ 5. What does the adversarial training \\ try to force the encoder representations of?\end{tabular}} \\ \hline
\end{tabular}
\caption{Human-annotated questions V.S. model-generated questions of three scientific papers from the SciMRC dataset. In general, human-annotated questions show a higher difficulty level.}
\label{tab:qg_example}
\end{table*}

\begin{table*}[t]
\centering
\begin{tabular}{p{0.2\textwidth}|p{0.7\textwidth}}
\hline
\textbf{Initial Summary} & \textbf{S0:} This paper proposes a new transfer learning approach for NMT that uses cross-lingual pre-training to enable a high performance on zero-shot translation ...... The proposed approach outperforms very strong baseline systems, consistently surpassing pivoting and multilingual approaches. Meanwhile, the performance on supervised translation direction remains the same level or even better when using their method. \\
\hline
\textbf{Feedback} & \textbf{Q:} Does the approach proposed by the authors perform better than all the comparison systems? \\ 
        & \textbf{F:} \textcolor{ForestGreen}{The proposed approach outperforms very strong baseline systems} \\
\hline
\textbf{Subsequent} \\ \textbf{Summaries} & \textbf{S1:} The paper proposes an approach to training a Chinese NER system-annotated data using adversarial training …… the negative influences of different crowd annotators. \textcolor{ForestGreen}{The proposed model outperforms very strong baseline systems.} \\
& \textbf{S2:} The paper proposes an approach to training a Chinese NER system on crowd-annotated data using adversarial training …… by reducing the negative influences of different crow annotators. \textcolor{ForestGreen}{The proposed model outperforms very strong baseline systems.} \\
& \textbf{S3:} The paper proposes an approach to training a Chinese NER system on crow-annotated data using …… \textcolor{ForestGreen}{The proposed approach outperforms very strong baseline systems.} The resulting features of the common Bi-LSTM are worker-invariant and NER-sensitive. \\
\hline
\textbf{Initial Summary} & \textbf{S0:} This paper explores transfer learning in a common zero-shot scenario where there are a lot of source-pivot and pivot-target parallel data but no source-target parallel data. The authors propose a new transfer learning approach for NMT that uses cross-lingual pre-training to enable a high performance on zero-shot translation ...... In addition, the performance on supervised translation direction remains the same level or even better when using their method. \\
\hline
\textbf{Feedback} & \textbf{Q:} What new transfer learning approach do authors propose for NMT? \\ 
        & \textbf{F:} \textcolor{ForestGreen}{We propose a new transfer learning approach for NMT that uses cross-lingual language model pre-training to enable high performance on zero-shot translation.} \\
\hline
\textbf{Subsequent} \\ \textbf{Summaries} & \textbf{S1:} The proposed approach uses \textcolor{ForestGreen}{cross-lingual} pre-training to alleviate the distance between different source language spaces and significantly \textcolor{ForestGreen}{improves zero-shot translation performance}, consistently surpassing pivoting and multilingual approaches. \\
& \textbf{S2:} The paper proposes \textcolor{ForestGreen}{a new transfer learning approach for NMT that uses cross-lingual language model pre-training to enable high performance on zero-shot translation}. The proposed approach uses cross-lingual pre-training to alleviate the distance between different source language...... \\
& \textbf{S3:} The main contributions of this paper are:* \textcolor{ForestGreen}{Proposing a new transfer learning approach for NMT that uses cross-lingual language model pre-training to enable high performance on zero-shot translation.} ...... Demonstrating that the proposed approach significantly \textcolor{ForestGreen}{improves zero-shot translation performance}, consistently surpassing pivoting and multilingual approaches.
 \\
\hline
\end{tabular}
\caption{Cases of LLMs \textbf{memorizing} \textcolor{ForestGreen}{facts} provided in the feedback and appear in subsequent summaries.}
\label{tab:fact}
\end{table*}

\begin{table*}[t]
\centering
\begin{tabular}{p{0.2\textwidth}|p{0.7\textwidth}}
\hline
\textbf{Feedback} & \textbf{Q:} What is the architecture of the framework \\ & \textbf{F:}  \textcolor{Maroon}{In this paper, we design a generation-based model, which is more efficient than the previous approach.}\\
\hline
\multicolumn{1}{l|}{\begin{tabular}[c]{@{}l@{}}\textbf{Subsequent}\\ \textbf{Summary}\end{tabular}} & \textbf{S:} The paper presents a new dataset for knowledge-based machine reading comprehension, which requires reasoning over triplet-style knowledge involved in a document. The authors find that published datasets do not sufficiently support this task, so they build a new dataset consisting of 40,047 examples for the knowledge-based machine reading comprehension task. \textcolor{ForestGreen}{They also propose a framework consisting of a question-answering model and a question-generation model}, both of which take the knowledge extracted from the document as well as relevant facts from an external knowledge base such as Freebase/ProBase/Reverb/NELL......The authors also implement two question-answering models, which directly measure the semantic similarity between \\
\hline
\textbf{Original Paper} & \textbf{P:} ...... The annotation of this dataset is designed so that successfully answering the questions requires understanding and the knowledge involved in a document. Each instance is composed of a question, a set of triplets derived from a document, and the answer. \textcolor{ForestGreen}{We implement a framework consisting of both a question-answering model and a question-generation model}, both of which take the knowledge extracted from the document as well as relevant facts from an external knowledge base such as Freebase/ProBase/Reverb/NELL ...... \\
\hline
\textbf{Feedback} & \textbf{Q:} What is the main purpose of collecting the data set? \\ & \textbf{F:} \textcolor{Maroon}{The main purpose of collecting the data set was to create a knowledge base for a dialog system.} \\ 
\hline
\multicolumn{1}{l|}{\begin{tabular}[c]{@{}l@{}}\textbf{Subsequent}\\ \textbf{Summary}\end{tabular}} & \textbf{S:} The paper proposes a dataset for developing and evaluating dialog systems based on knowledge bases. The proposed dataset aims to help improve the accuracy of dialog systems in providing general information from a knowledge base...... The authors hypothesize that users are willing to give such explanations to help find answers for their own questions. \textcolor{ForestGreen}{The main purpose of the collected dataset is to enable interactive learning using the proposed approach and potentially to evaluate how different systems perform on this task.} The paper also distinguishes three different basic approaches to requesting knowledge bases.\\
\hline
\textbf{Original Paper} & \textbf{P:} ...... The last source of information that we consider for interactive learning is rephrasing, which could help when the system does know the concept but does not know the correct wording. This area is extensively studied for the purposes of information retrieval BIBREF5, BIBREF6. \textcolor{ForestGreen}{The main purpose of the collected dataset is to enable interactive learning using the steps proposed above and potentially to evaluate how different systems perform on this task.} The obvious difficulty when developing a dialog system is finding a way how to identify the piece of information that the user is interested in ...... \\
\hline
\end{tabular}
\caption{Cases of LLMs being \textbf{capable of rectifying} \textcolor{Maroon}{non-facts} provided in the feedback into \textcolor{ForestGreen}{facts} in the subsequent summary. The source paper is presented as a reference to verify the facts.}
\label{tab:nonfact-success}
\end{table*}

\begin{table*}[t]
\centering
\begin{tabular}{p{0.2\textwidth}|p{0.7\textwidth}}
\hline
\textbf{Feedback} & \textbf{Q:} What is the limitation of the current state-of-the-art in NMT? \\ & \textbf{F:}  \textcolor{Maroon}{The current state-of-the-art in Neural Machine Translation (NMT) heavily relies on large-scale parallel data, resulting in poor performance on low-resource or zero-resource language pairs.}\\
\hline
\multicolumn{1}{l|}{\begin{tabular}[c]{@{}l@{}}\textbf{Subsequent}\\ \textbf{Summary}\end{tabular}} & \textbf{S:} \textcolor{Maroon}{The current state-of-the-art in Neural Machine Translation (NMT) heavily relies on large-scale parallel data, resulting in poor performance on low-resource or zero-resource language pairs.} One common alternative to avoid pivoting in NMT is transfer learning, which leverages a high-resource pivot-target model to initialize a low-resource source-target model that is further optimized with a small amount of available parallel data...... \\
\hline
\textbf{Feedback} & \textbf{Q:} How to create synthesized summaries in this paper? \\ & \textbf{F:} \textcolor{Maroon}{We propose a template-based synthesis method to create synthesized summaries, then explore the effect of enriching training data for abstractive summarization using the proposed model compared to a synthesis baseline} \\ 
\hline
\multicolumn{1}{l|}{\begin{tabular}[c]{@{}l@{}}\textbf{Subsequent}\\ \textbf{Summary}\end{tabular}} & \textbf{S:} \textcolor{Maroon}{1. We use a simple template-based method to create synthesized summaries.} 2. We propose a novel abstractive summarization model based on the encoder-decoder architecture, which achieves state-of-the-art performance. 3. We provide a new summarization dataset for student reflections. 4. We provide a baseline of a seq2seq model. 5. In domains such as student reflections, the lack of sufficient training data may be a problem. 6. We explore domain transfer for abstractive summarization.\\
\hline
\textbf{Feedback} & \textbf{Q:} What do they plan to do in the future? \\ & \textbf{F:} \textcolor{Maroon}{In the future, we plan to do the following: Explore the application of NMT in multilingual learning} \\
\hline
\multicolumn{1}{l|}{\begin{tabular}[c]{@{}l@{}}\textbf{Subsequent}\\ \textbf{Summary}\end{tabular}} & \textbf{S:} The paper introduces a unified approach to seamlessly extend the original NMT to multilingual scenarios. The proposed framework allows any language to be integrated into any side of the encoder-decoder architecture with only one encoder and one decoder for all the languages involved...... and the results show that bringing multilingualism to NMT helps to improve individual translations. \textcolor{Maroon}{In the future, the authors plan to explore the application of NMT in multilingual learning} and evaluate the proposed framework on more languages and more translation scenarios. \\
\hline
\end{tabular}
\caption{Cases of LLMs \textbf{failing to exclude or rectify} \textcolor{Maroon}{non-facts} provided in the feedback in the subsequent summary, given the \textcolor{Maroon}{non-facts} also appear in the subsequent summaries.}
\label{tab:nonfact-fail}
\end{table*}

\section{Prompts}
\label{sec:prompt}
We show all our prompt templates below.
\subsection{Baseline prompts}
\label{sec:bpp}
\paragraph{Prompt 1}
\texttt{\\ Read this: ``This is the body text of a scientific paper'' \\ Write a summary.}

\paragraph{Prompt 2}
\texttt{\\ Below is a scientific paper. Please write a summary about it.\\  Paper: ``This is the body text of a scientific paper''}

\subsection{Pre-hoc prompt engineering}
\label{sec:ppe}
\paragraph{Prompt 1}
\texttt{\\ Below is a scientific paper. Please write a summary that is factually consistent with the paper.\\ Paper: ``This is the body text of a scientific paper''}

\paragraph{Prompt 2}
\texttt{\\ Write a summary of the scientific paper below. Reduce non-facts as much as possible in the summary you generate.\\ Paper: ``This is the body text of a scientific paper''}

\paragraph{Prompt 3}
\texttt{\\ Please write a summary about the scientific paper below. Make sure the summary is factually consistent with the paper. Do not include non-factual information. \\ Paper: ``This is the body text of a scientific paper''}

\subsection{Feedback generation}
\label{sec:fgp}
\paragraph{\feedback{}}
\texttt{\\ Below is a question paired with context. Please return your response in two parts: \\1. The answer to the question\\2. The most relevant evidence sentence in the context to your answer.\\ If the question is unanswerable, directly return ‘Unanswerable’.\\ Question: ``This is the body text of a question'' \\ Context: ``This is the body text of a scientific paper''}
\paragraph{Generic feedback}
\texttt{\\ Below is a scientific paper paired with feedback. Please write a summary based on the feedback provided.\\ Paper: ``This is the body text of a scientific paper'' \\ Feedback: ``The previous summary you generated is not factually consistent with the paper. Please re-generate a summary with higher factuality. Reduce your factual errors and include more facts.''}

\subsection{Summary generation with feedback}
\texttt{\\ Below is a scientific paper paired with feedback on factual information. Please write a summary by memorizing the provided facts and rectifying the provided non-facts.\\ Facts:\{$sent_1, sent_2, \cdots, sent_n$\}\\
Non-Facts:\{$sent_1, sent_2, \cdots, sent_n$\}\\ Paper: ``This is the body text of a scientific paper.''}

\subsection{GPT-3.5}
\texttt{\\ SYSTEM: You are good at answering reading comprehension questions and extracting corresponding evidence. \\ USER: Read the context\\ Context: ``This is the body of a context'' \\ Please answer the following question by:\\ 1. Provide the answer\\ 2. Provide the evidence sentence in the context.\\ Please use “Answer:” and “Evidence:” to denote these two parts when generating your response: ``This is the body of a question''}

\section{Cases}
\label{sec:cases}
In Table \ref{tab:fact}, we provide cases of LLMs' behaviors when confronted with facts in the feedback. In Table \ref{tab:nonfact-success}, we provide successful cases of LLMs' behaviors when confronted with non-facts in the feedback. In Table \ref{tab:nonfact-fail}, we provide error cases of LLMs' behaviors when confronted with non-facts in the feedback.

\end{document}